\documentclass[11pt]{article}
\usepackage{smile}
\usepackage{biblatex}
\usepackage[colorlinks,
           linkcolor=blue,
           anchorcolor=blue,
           citecolor=blue
           ]{hyperref}

\usepackage{kpfonts}
\DeclareMathAlphabet{\mathsf}{OT1}{cmss}{m}{n}
\SetMathAlphabet{\mathsf}{bold}{OT1}{cmss}{bx}{n}

\usepackage{fullpage}

\usepackage{amsmath, amssymb, amsfonts}
\usepackage{bm}
\usepackage{algorithm, algorithmic}
\usepackage{graphics}
\usepackage{graphicx}
\usepackage{wrapfig}
\usepackage{svg}
\usepackage{xcolor}
\usepackage[font=footnotesize]{caption}
\usepackage{comment}

\newcommand*{\ZhigenZ}{\textcolor{black}}

\graphicspath{{./figures/}}

\author{
 Zhigen Zhao\thanks{School of Mechanical Engineering, Georgia Institute of Technology}, 
 Simiao Zuo\thanks{School of Industrial and Systems Engineering, Georgia Institute of Technology}, 
 Tuo Zhao\footnotemark[2] \thanks{Co-corresponding Author}, 
 Ye Zhao\footnotemark[1] \footnotemark[3]\\
\texttt{\{zhigen.zhao, simiaozuo, tourzhao, yzhao301\}@gatech.edu}
}

\begin{document}

\title{\huge Adversarially Regularized Policy Learning Guided by Trajectory Optimization}

\date{}
\maketitle
\vspace{-0.3in}
\begin{abstract}
Recent advancement in combining trajectory optimization with function approximation (especially neural networks) shows promise in learning complex control policies for diverse tasks in robot systems. Despite their great flexibility, the large neural networks for parameterizing control policies impose significant challenges. 
The learned neural control policies are often overcomplex and non-smooth, which can easily cause unexpected or diverging robot motions.
To address this issue, we propose ad\textbf{ve}rsarially \textbf{r}egularized p\textbf{o}licy lear\textbf{ni}ng guided by traje\textbf{c}tory optimiz\textbf{a}tion (VERONICA) for learning smooth control policies. Specifically, our proposed approach controls the smoothness (local Lipschitz continuity) of the neural control policies by stabilizing the output control with respect to the worst-case perturbation to the input state. Our experiments on robot manipulation show that our proposed approach not only improves the sample efficiency of neural policy learning but also enhances the robustness of the policy against various types of disturbances, including sensor noise, environmental uncertainty, and model mismatch.

\end{abstract}

\section{Introduction}
Robust and generalizable motion planning enables robotic systems to handle various uncertainties and accomplishes diverse tasks. However, learning a dynamically consistent neural control policy (i.e., a neural-network control policy)  and executing it reliably remain challenging. First, the function approximators used to model the policy can be highly complex and non-smooth, causing poor generalization performance. \ZhigenZ{Second, the dynamics models involved are often mismatched from the physical robot, leading to the need 
of learning a robust policy.}





Trajectory optimization (TO)~\cite{betts1998survey, kuindersma2016optimization,  tassa2014control, posa2014direct} is a powerful model-based approach to generate optimal control sequences for complex robotic systems. However, existing methods for solving TO problems with full robot dynamics require solving large nonlinear programs, resulting in high computational cost. This difficulty prevents the use of TO methods in real-time robot control settings.
As such, to alleviate the computational burden at run-time, it is preferable to have a parametric representation of a robot control policy. 
In comparison, model-free policy search, as in~\cite{deisenroth2013survey}, aims to automatically learn the controller through random exploration. However, a majority of these methods fail to \ZhigenZ{utilize the prior knowledge on the robot dynamics encoded in the physical model}, which causes sample inefficiency.



To take advantage of both TO and policy search,
\cite{mordatch2014combining} and \cite{levine2013guided} train a robot control policy supervised by optimized trajectory samples, and meanwhile adapting TO to the learned policy. The work in
\cite{mordatch2014combining} observes that the derivatives of a neural control policy can behave irregularly even when the policy matches the optimal trajectory baseline. This is because neural networks have high complexity and flexibility, which makes them highly non-smooth --- a small change in the networks' input can cause a large variation in the output. To mitigate this limitation, existing works attempt to impose some smoothness constraints on the policy. For instance, \cite{mordatch2014combining} matches the gradient for policy and trajectory samples via tangent propagation. However, tangent propagation requires Jacobian computation on each trajectory point, which does not scale well to large datasets.

To alleviate these issues, we propose a new approach: ad\textbf{ve}rsarially \textbf{r}egularized p\textbf{o}licy lear\textbf{ni}ng guided by traje\textbf{c}tory optimiz\textbf{a}tion (VERONICA). Specifically, our approach improves the local Lipschitz continuity of the neural control policy via adversarial regularization, which improves generalization performance for inputs not seen during training. We focus on promoting smoothness in policy for non-hybrid robotics tasks that are often governed by differential equations with high-order continuity. 
For hybrid systems where non-smooth dynamics might occur during physical contact, several works in TO \cite{brubaker2009estimating, todorov2011convex, mordatch2012discovery} and physical simulation MuJoCo \cite{todorov2012mujoco} propose to model contact with a smoothed model, where contact forces diminish gradually with contact distance. 
The work of \cite{drnach2021robust} proposes a risk-sensitive cost function to represent a stochastic, smoothed variant of the original complementarity contact problem \cite{chen1996class}. In this work, we show that the VERONICA framework also provides robustness benefits for a hybrid locomotion system with physical contacts.

The VERONICA framework is related to existing works \cite{miyato2018virtual,zhang2019theoretically,hendrycks2019using,xie2019unsupervised,jiang2019smart,shen2020deep, jiang2019smart, zuo2021arch, li2020implicit}. These works consider similar regularization techniques, but target at other applications with different motivations, e.g., semi-supervised learning, unsupervised domain adaptation,  harnessing adversarial examples, fine-tuning pre-trained models and reinforcement learning. \ZhigenZ{ \cite{morimoto2000robust} and \cite{pinto2017robust} solve similar min-max problems to improve the robustness of reinforcement learning.}

We further observe that besides promoting policy smoothness, adversarial regularization improves the robustness of the policy against modeling errors and perturbations in the environment. We verify that the VERONICA framework produces stable robot behaviors under sensor noise, environmental uncertainty, and model mismatch.

Conventionally, adversarial regularization involves a min-max game, which is solved by alternating gradient descent-ascent. During training, neither of the players can be advantageous, such that the generated perturbations can be over-strong and hinder model generalization.
To resolve this issue, we employ Stackelberg adversarial regularization (SAR), as proposed in~\cite{zuo2021adversarial}, which formulates adversarial regularization as a Stackelberg game~\cite{von2010market}. In SAR, the policy (i.e., the leader) has a higher priority than the perturbation (i.e., the follower). The leader procures its advantage by considering how the follower will respond after observing the leader’s decision, such that the leader anticipates the predicted move of the follower when optimizing its strategy.
We note that prioritizing the policy optimization is reasonable and beneficial because we target the performance of the learned policy, instead of the adversary.


Our contributions are: I) We propose VERONICA, an adversarial regularization method for learning smooth neural control policies guided by TO. This improves the generalization performance of the learned policy; 
II) We show that the learned policy achieves better robustness under disturbances such as sensor noise, environmental uncertainty, and model mismatch; III) We reformulate adversarial regularization as a Stackelberg game, which further improves generalization and robustness of the policy compared with the conventional formulation.

\section{Related Works}
\textbf{Adversarial Training in Robot Learning:} Adversarial training has previously been used to improve safety in robot visuomotor control scenarios \cite{chen2020adversarial}. The work in \cite{lechner2021adversarial} argues that adversarial training induces unexplored error profiles in \textit{vision-based} robot learning, which studies classification tasks that are not Lipschitz continuous. In contrast, our work focuses on adversarial regularization for neural control policy in \textit{dynamics-based} robot learning, which are intrinsically smooth. Therefore, \textit{vision-based} adversarial training studies fundamentally different problems than ours.



\textbf{Imitation Learning: }
Behavioral cloning (BC) uses supervised learning to directly imitate expert trajectories without interacting with the environment \cite{schaal1997learning}. However, BC is particularly vulnerable to error compounding \cite{ross2011reduction}. In our work, we solve a BC problem for policy learning in each iteration of the Alternating Direction Method of Multipliers (ADMM) method, while the ADMM framework offers a coupling mechanism to allow the trajectory optimizer (i.e., the teacher) to not only guide the learned policy (i.e., the student) towards better solutions but also adapt to the student. More importantly, we incorporate an adversarial regularizer to improve policy smoothness, which significantly eases the effect of error compounding.

Along another line of research, 
generative adversarial imitation learning \cite{ho2016generative, zolna2019task} uses generative adversarial networks (GAN) to directly generate policies that imitate expert demonstrations. In contrast, the adversaries in our work are the direct perturbations on the input (i.e., the robot state), rather than the discriminator network.

\textbf{Trajectory-Optimization-Guided Policy Learning:} Trajectory optimization has been used to aid and stabilize value function learning in the reinforcement learning (RL) context \cite{lowrey2018plan}, while the authors of \cite{landry2021SEAGuL} use a bilevel optimization to learn the value function with adversarial samples. In this work, we focus on supervised learning approaches that train neural control policies from TO.

Guided policy search (GPS)~\cite{levine2013guided, levine2013variational, levine2014learning} iteratively updates guiding sample using differential dynamic programming (DDP) and trains policies on the distribution over the guiding samples. 
In contrast, the work of \cite{mordatch2014combining} seeks consensus between neural network policy and trajectory optimization using ADMM \cite{boyd2011distributed}. The authors in \cite{duburcq2020online} similarly solve for ADMM consensus, but aim to learn a trajectory sequence rather than policy. The ADMM formulation in our work is closely related to \cite{mordatch2014combining}, but we focus on adversarial regularization for policy learning. 

\section{Method}
We introduce VERONICA, our proposed adversarially regularized approach which combines the strength of policy learning and trajectory optimization. First, we define an adversarial regularizer and explain how it improves smoothness and robustness of neural control policies; Second, we describe an ADMM-based algorithm that solves the full joint optimization problem; Third, we develop an extension to our proposed adversarial regularization approach --- Stackelberg adversarial regularization.
We consider the neural control policy learning process guided by $N$ optimal trajectories $\{\mathbf{X},\mathbf{U}\}=\{\mathbf{X}_i, \mathbf{U}_i~|~i=1,\cdots,N\}$, and each optimal trajectory $\{\mathbf{X}_{i}, \mathbf{U}_{i}\}$ consists of $T$ state-control pairs $\{\mathbf{x}_i^t\in\mathbb{R}^{d_x}, \mathbf{u}_i^t\in\mathbb{R}^{d_u}~|~t=1,\cdots,T\}$, where $\mathbf{x}_{i}^{t}$ and $\mathbf{u}_{i}^{t}$ denote the robot state and the control, respectively. In this study, the robot state corresponds to the joint positions, velocities and task parameters such as goal configurations, while the control corresponds to the joint torque. Moreover, let $\pi(\cdot|\textbf{W})$ denotes the neural control policy, where $\textbf{W}$ denotes the associated parameters.

\subsection{Adversarial Regularization for Neural Control Policy} \label{sec:adv_learning}


To promote smoothness of the neural control policy, we consider the following adversarial discrepancy measure:
\begin{align*}
    r_{\rm adv}(\mathbf{x}, \mathbf{W})
    =\max_{\|\delta\|\leq\epsilon}r(\mathbf{x},\mathbf{W},\boldsymbol{\delta})
    =\max_{\|\delta\|\leq\epsilon}\|\pi(\mathbf{x}|\mathbf{W})-\pi(\mathbf{x}+\boldsymbol\delta|\mathbf{W})\|^2,
\end{align*}
where $\|\cdot\|$ denotes the $\ell_2$ norm, $\boldsymbol\delta\in\mathbb{R}^{d_x}$ is the adversarial perturbation injected to the state vector $\mathbf{x}$, and $\epsilon >0$ is the perturbation strength. Such an adversarial discrepancy measure $r_{\rm adv}(\mathbf{x}, \mathbf{W})$ essentially computes the maximal deviation of the neural control policy output at state $\mathbf{x}$ given an input perturbation $\boldsymbol \delta$ whose $\ell_2$ norm is bounded by $\epsilon$.

We then apply the adversarial discrepancy measure to control the smoothness of the neural control policy. Specifically, we solve the following joint optimization problem:
\ZhigenZ{
\begin{align}
    \min_{\mathbf{X}, \mathbf{U}, \mathbf{W}}\sum_{i=1}^{N}&\mathcal{L}(\mathbf{X}_{i}, \mathbf{U}_{i}) + \mathcal{Q}_{\rm BC}(\mathbf{X},\mathbf{U},\mathbf{W}) + \alpha\mathcal{R}_{\rm adv}(\mathbf{X},\mathbf{W}),\label{eqn:joint_opt}\\
    \nonumber\text{s.t.}\quad\mathbf{x}^{t+1}&=f(\mathbf{x}^{t}, \mathbf{u}^{t}), 
    \mathbf{x}^0=\mathbf{x}_{\rm{init}}, \mathbf{X} \in \mathcal{X}, \mathbf{U} \in \mathcal{U},
\end{align}}
where $\mathcal{L}(\mathbf{X}_i, \mathbf{U}_i)$ denotes the loss function of the trajectory optimization (TO) for the $i^{\rm th}$ trajectory, $\mathcal{Q}_{\rm BC}(\mathbf{X},\mathbf{U},\mathbf{W})$ denotes the loss function for policy learning:
\begin{align*}
    \mathcal{Q}_{\rm BC}(\mathbf{X}, \mathbf{U}, \mathbf{W}) = \frac{1}{N}\sum_{i, t}||\pi(\mathbf{x}_i^t|\mathbf{W})-\mathbf{u}_i^t||^2,
\end{align*}
$\mathcal{R}_{\rm adv}(\mathbf{X},\mathbf{W})$ is the adversarial regularizer for controlling the smoothness of the policy:
\begin{align*}
    \mathcal{R}_{\rm adv}(\mathbf{X},\mathbf{W}) = \frac{1}{N}\sum_{i,t}r_{\rm adv}(\mathbf{x}_i^t, \mathbf{W})=\frac{1}{N}\sum_{i,t}\max_{\|\boldsymbol{\delta_{i}^t}\|\leq\epsilon}\|\pi(\mathbf{x}_i^{t}|\mathbf{W})-\pi(\mathbf{x}_{i}^t+\boldsymbol\delta_{i}^t|\mathbf{W})\|^2,
\end{align*}
and $\alpha$ is the regularization coefficient weighting between the $\mathcal{Q}_{\rm BC}(\mathbf{X}, \mathbf{U}, \mathbf{W})$ and $\mathcal{R}_{\rm adv}$.

Solving the optimization problem in Eq.~{\eqref{eqn:joint_opt}} learns a neural control policy that not only minimizes the TO loss and the behavior cloning loss, but also encourages the adversarial discrepancy measure of the policy to be small at every state of the optimal trajectories. 

\begin{figure}[htb!]
\centering
\includegraphics[width=0.4\textwidth]{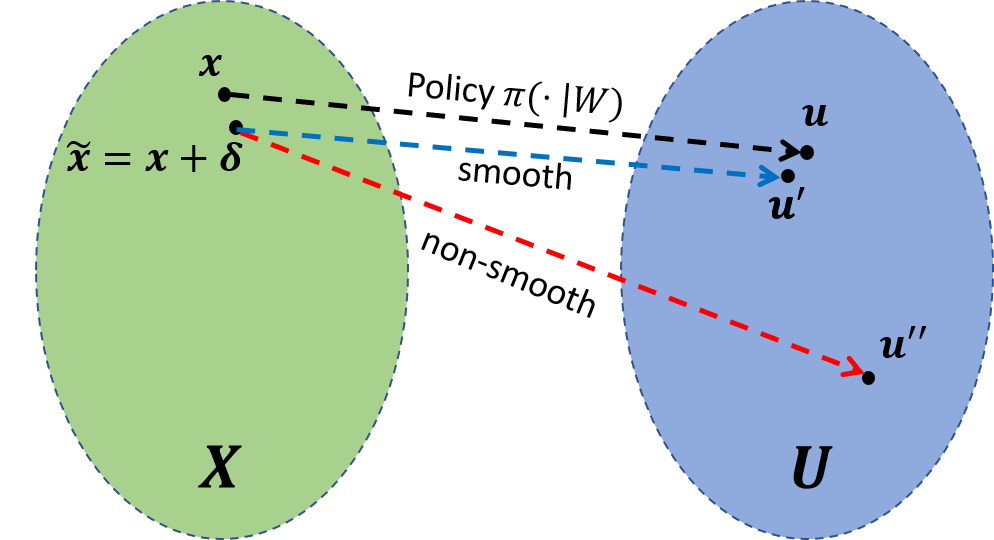}
\caption{Illustration of policy smoothness at state $\mathbf{x}$ and control $\mathbf{u}$. If the policy $\pi(\cdot|W)$ is smooth around $\mathbf{x}$, the perturbed state $\mathbf{\Tilde{x}}$ will produce a control $\mathbf{u'}$ similar to $u$. If the policy $\pi(\cdot|\mathbf{W})$ is non-smooth around $\mathbf{x}$, the output control $\mathbf{u''}$ would deviate significantly from $\mathbf{u}$.}
\label{fig:smoothness_illustration}
\end{figure}

\textbf{(I) Adversarial Regularization Improves Generalization:}
Existing methods usually train neural control policies by only minimizing the trajectory optimization loss and behavior cloning loss. Due to the high capacity of deep neural networks, the learned neural control policies are often over-complex and highly non-smooth. This is inconsistent with observations that many optimal control policies for robots are smooth. Here we exclude the problem involving physical contact dynamics, which exhibits discontinuous and non-smooth phenomenon.
Smoothness requires a small perturbation to the state vector $\mathbf{x}$ to only yield a small change to the policy output (Figure~\ref{fig:smoothness_illustration}).
\ZhigenZ{Such a property is desirable in robotics tasks, since they often involve differential equations with high-order continuity properties. Therefore, improving smoothness of the learned policy can improve its ability to generalize to states unseen during training.}

VERONICA naturally promotes the desired smoothness by imposing a high penalty when the adversarial perturbation $\boldsymbol \delta$ yields a large deviation to the policy output. More precisely, $r_{\rm adv}(\mathbf{x},\mathbf{W})$ essentially upper bounds the deviation of the policy output due to the adversarial perturbation $\boldsymbol\delta$ with respect to the state $\mathbf{x}$, and therefore
can be viewed as a measure of the local Lipschitz constant within a small neighborhood of $\mathbf{x}$, 
i.e., $C_{\mathbf{x}}=\sup_{\|\boldsymbol\delta\|\leq\epsilon}\frac{\|\pi(\mathbf{x}|\mathbf{W})-\pi(\mathbf{x}+\boldsymbol{\delta}|\mathbf{W})\|}{\|\boldsymbol{\delta}\|}.$
Accordingly, our proposed adversarial regularizer penalizes the average discrepancy measures of the neural control policy at all trajectory points, which enforces its local Lipschitz continuity.

\begin{remark}
An alternative regularization technique is the so-called Jacobian regularization (JR), which penalizes the Frobenius norm of the Jacobian matrix of the policy with respect to the input state at all trajectory points $\mathcal{R}_{\rm JR}(\mathbf{X},\mathbf{W}) = \frac{1}{N}\sum_{i,t}\|\nabla_{\mathbf{x}}\pi(\mathbf{x}_i^{t}|\mathbf{W})\|_{\rm F}^2$.
As shown in \cite{yang2020adversarial}, such a Jacobian regularizer is not particularly effective in promoting the Lipschitz continuity of large neural networks. Moreover, when using the Jacobian regularizer for stochastic gradient type algorithms, one needs to further differentiate through the Jacobian with respect to the parameter $\mathbf{W}$, which is neither computationally efficient nor scalable in practice.
\end{remark}

\textbf{(II) Adversarial Regularization Gains Robustness:} Robot systems measure their states from sensors, which are prone to stochastic or systematic sensor errors.
VERONICA naturally gains robustness against such disturbances. Specifically, the adversarial perturbation in VERONICA can be viewed as a proxy to the errors. Therefore, our approach does not require prior knowledge of them. In comparison, existing methods for handling such errors usually assume specific forms, e.g., independent Gaussian noise, which can be restrictive in practice.

Moreover, as suggested in \cite{asadi2018lipschitz}, the Lipschitz continuity is essential to robustness, especially for control and reinforcement learning problems. This is because for policies without the Lipschitz continuity property, a small error in sensor measurement or state transition potentially leads to a drastic change to the policy output. Due to the dynamic nature of the control problem, it will further yield significant error compounding during policy roll-out. 
Moreover, when the models used to describe robot dynamics mismatch the real robot, such compounding system errors can be catastrophic. 
\ZhigenZ{Quantitatively, the upper bound for policy robustness under state disturbance, measured by compounding value function discrepancy, is proportional to the Lipschitz constant of the neural control policy (Appendix \ref{supp_sec:robustness}).} As the VERONICA approach can effectively control the local Lipschitz continuity of the neural control policy, such an issue can be mitigated.

\subsection{Combined Trajectory Optimization and Adversarially Regularized Policy Learning}
\label{sec:combined_learning_and_to}

We apply ADMM~\cite{Zhao2020SyDeBO, Ziyi2020ADMM} to solve the optimization problem in Eq.~{\eqref{eqn:joint_opt}}. Specifically, we reparameterize Eq.~{\eqref{eqn:joint_opt}} into a decomposable form by introducing two auxiliary sets of state and control variables: $(\mathbf{X}^{\rm TO}, \mathbf{U}^{\rm TO})$ represents the trajectory samples generated by trajectory optimization (TO), and $(\mathbf{X}^{\rm PL}, \mathbf{U}^{\rm PL})$ are copies of $(\mathbf{X}^{\rm TO}, \mathbf{U}^{\rm TO})$ for policy learning. Accordingly, the optimization problem in Eq.~{\eqref{eqn:joint_opt}} is reformulated as:
\begin{align}
        \min_{\substack{\mathbf{X}^{\rm TO, PL},\mathbf{U}^{\rm TO, PL},\mathbf{W}}} ~~ & \sum_{i=1}^{N}\mathcal{L}(\mathbf{X}^{\rm TO}_{i}, \mathbf{U}^{\rm TO}_{i})+ \mathcal{Q}_{\rm BC}(\mathbf{X}^{\rm PL},\mathbf{U}^{\rm PL},\mathbf{W})+ \alpha\mathcal{R}_{\rm adv}(\mathbf{X}^{\rm PL},\mathbf{W})\label{eqn:admm_joint_opt} \notag \\
        & \textrm{s.t.} \hspace{0.08in} \mathbf{X}^{\rm TO}=\mathbf{X}^{\rm PL}, \mathbf{U}^{\rm TO}=\mathbf{U}^{\rm PL}.
\end{align}
ADMM splits the above optimization problem into $N$ individual TO problems and a policy learning problem to be solved in an iterative manner. Let $\boldsymbol{\lambda}^p_{\mathbf{X}_i}, \boldsymbol{\lambda}^p_{\mathbf{U}_i}$ denote the dual variables at the $p^{\rm th}$ iteration and $\rho_x, \rho_u >0$ denote the penalty parameters. The ADMM primal and policy updates are:

\begin{eqnarray}
\hspace{-0.5in}&&\mathbf{X}_i^{{\rm TO},p+1},  \mathbf{U}_i^{{\rm TO},p+1} = \;   \underset{\mathbf{X}_i, \mathbf{U}_i}{\arg\min}\;\;  \mathcal{L}(\mathbf{X}_i, \mathbf{U}_i) + \frac{\rho_x}{2}\|\mathbf{X}_i-\mathbf{X}_i^{{\rm PL},p}+\boldsymbol{\lambda}_{\mathbf{X}_i}^{p}\|^2\notag\\
\hspace{-0.5in}&& \hspace{1.85in}+ \frac{\rho_u}{2}\|\mathbf{U}_i-\mathbf{U}_i^{{\rm PL},p}+\boldsymbol{\lambda}_{\mathbf{U}_i}^{p}\|^2, \quad \mbox{(\textbf{primal TO update})} \label{eqn:admm_to}\\
 \hspace{-0.5in} && \mathbf{W}^{p+1} = \;   \underset{\mathbf{W}}{\arg\min} \;\;\mathcal{Q}_{\rm BC}(\mathbf{X}^{{\rm PL},p},\mathbf{U}^{{\rm PL},p},\mathbf{W}) + \mathcal{R}_{\rm adv}(\mathbf{X}^{{\rm PL},p},\mathbf{W}), \quad  \mbox{(\textbf{policy update})} \label{eqn:admm_pl} \\
\hspace{-0.5in} &&  \mathbf{X}_i^{{\rm PL},p+1},  \mathbf{U}_i^{{\rm PL},p+1} = \;  \underset{\mathbf{X}_i, \mathbf{U}_i}{\arg\min}\;\; \mathcal{Q}_{\rm BC}(\mathbf{X}_i^{{\rm PL},p}, \mathbf{U}_i^{{\rm PL},p}, \mathbf{W}^{p+1}) + \frac{\rho_x}{2}\|\mathbf{X}_i^{{\rm TO},p+1}-\mathbf{X}_i+\boldsymbol{\lambda}_{\mathbf{X}_i}^{p}\|^2 \nonumber\\
\hspace{-0.5in} & & \hspace{1.85in}+ \frac{\rho_u}{2}\|\mathbf{U}_i^{{\rm TO},p+1}-\mathbf{U}_i+\boldsymbol{\lambda}_{\mathbf{U}_i}^{p}\|^2. \quad \mbox{(\textbf{primal PL update})} \label{eqn:xr_update_main}
\end{eqnarray}

\textbf{Primal TO update:}
The update in Eq.~{\eqref{eqn:admm_to} involves TO and is solved by either direct optimization methods or indirect methods such as differential dynamic programming (DDP)~\cite{tassa2014control, jacobson1970differential}. We defer details of the DDP algorithm to Appendix~\ref{appendix:DDP}.

\textbf{Policy update:}
Note that Eq.~{\eqref{eqn:admm_pl}} is a min-max optimization problem. For notation simplicity, we omit the iteration index $p$, and we rewrite it as
\begin{align}\label{minmax-pl}
\mathbf{W} =  \underset{\mathbf{W}}{\arg\min}~\mathcal{Q}_{\rm BC}(\mathbf{X}^{\rm{PL}},\mathbf{U}^{\rm{PL}},\mathbf{W}) + \frac{\alpha}{N}\sum_{i,t}\max_{\|\boldsymbol{\delta}_{i}^{t}\|\leq\epsilon} r(\mathbf{x}_i^{{\rm PL},t},\mathbf{W},\boldsymbol{\delta}_{i}^{t}).
\end{align}
To solve Eq.~{\eqref{minmax-pl}}, we apply an alternating gradient descent/ascent algorithm. Specifically, at the $s^{\rm th}$ iteration, we first apply
the projected gradient ascent algorithm to update $\boldsymbol{\delta}_{i}^{t}$ for $K$ steps,
\begin{align*}
\boldsymbol{\delta}_{i}^{t,s} = {\boldsymbol \delta}_i^{t, s, K},
\text{ where }
\boldsymbol{\delta}_{i}^{t, s, k} = \Pi \left[ \boldsymbol{\delta}_{i}^{t, s, k-1} + \eta_\delta\nabla_{\delta}
r(\mathbf{x}_i^{{\rm PL},t},\mathbf{W}^{s}, \boldsymbol{\delta}_{i}^{t,s,k-1}) \right]
\text{ for } k=2,\cdots ,K.
\end{align*}
Here, $\boldsymbol{\delta}_{i}^{t,s,1}$ is randomly sampled from $\mathcal{N}(0, \sigma^2 \mathbb{I})$, $\Pi$ denotes projection to the $\ell_2$ ball with a radius $\epsilon$, and $\eta_{\delta}>0$ denotes the step size. Then we apply a gradient descent (or stochastic gradient descent) step to $\mathbf{W}$,
\begin{align} \label{eqn:gd_w}
\mathbf{W}^{s} = \mathbf{W}^{s-1} - \eta_W[\nabla_{\mathbf{W}}\mathcal{Q}_{\rm BC}(\mathbf{X}^{\rm{PL}},\mathbf{U}^{\rm{PL}},\mathbf{W}^s) + \frac{\alpha}{N}\sum_{i,t} \nabla_{\mathbf{W}}r(\mathbf{x}_i^{{\rm PL},t},\mathbf{W}^{s},\boldsymbol{\delta}_{i}^{t,s})].
\end{align}
\textbf{Primal PL update:}
The update in Eq.~{\eqref{eqn:xr_update_main}} solves an unconstrained differentiable optimization sub-problem, which can be efficiently solved for each trajectory using stochastic gradient descent.

\textbf{Dual update:}
After the above three updates, we perform the dual update as follows:
\begin{align}\label{eqn:dual_updates}
    \boldsymbol{\lambda}_{\mathbf{X}_i}^{p+1}=
    \boldsymbol{\lambda}_{\mathbf{X}_i}^{p} + \mathbf{X}_i^{{\rm TO},p+1} - \mathbf{X}_i^{{\rm PL},p+1}, \quad 
    \boldsymbol{\lambda}_{\mathbf{U}_i}^{p+1}=
    \boldsymbol{\lambda}_{\mathbf{U}_i}^{p} + \mathbf{U}_i^{{\rm TO},p+1} - \mathbf{U}_i^{{\rm PL},p+1}.
\end{align}

After a certain number of iterations of the above primal-dual policy updates, the joint optimization in Eq.~{\eqref{eqn:admm_joint_opt}} achieves a consensus and the primal and dual residuals meet the ADMM stopping criteria. The overall algorithm is summerized in Algorithm~\ref{pseudo:admm_policy_learning} in Appendix~\ref{appendix-ADMM}.

\subsection{Stackelberg Adversarial Regularization}
\label{sec:SAR}

One major limitation of the adversarial regularizer in Eq.~{\eqref{minmax-pl}} is that it solves a min-max-game-based optimization, where neither of the players can be advantageous. This is problematic because the adversarial player may generate over-strong perturbations that hinder generalization. To mitigate this issue, we employ Stackelberg adversarial regularization~\cite{zuo2021adversarial} to solve the policy update in Eq.~{\eqref{minmax-pl}} through a Stackelberg game formulation.
In a Stackelberg game, there are two players, a leader (the policy) and a follower (the perturbations). The leader acknowledges the strategy of the follower, such that it is always in an advantageous position. This effectively eliminates the over-strong perturbations.

\begin{algorithm}[htb!]
    \caption{Adversarially Regularized Policy Learning.}
    \label{pseudo:adv_policy_learning}
    \textbf{Input: }$\{\mathbf{X, U}\}$: trajectory samples; $E$: number of epochs; $K$: number of perturbation updates.
    \begin{algorithmic}
    
    \FOR{$\mathrm{epoch}=1,\cdots,E$}
        \FOR{$\{\mathbf{x}, \mathbf{u}\}\in\{\mathbf{X, U}\}$}
        \STATE \textrm{Initialize} $\mathbf{\boldsymbol\delta}^0 \sim \mathcal{N}(0, \sigma^2\mathbb{I})$\;
        \FOR{$k=1,\cdots,K$}
        \STATE \textrm{Compute }$\rm{d}\mathcal{R}_{\rm adv}/\rm{d}\boldsymbol\delta^{k-1}$\;
        \STATE $\mathbf{\boldsymbol\delta}^k\gets\mathrm{Optimizer}(\mathrm{d}\mathcal{R}_{\rm adv}/\mathrm{d}\boldsymbol\delta^{k-1})$\;
        
        \ENDFOR
        \STATE \textbf{\textit{Adv Reg}: }
        \STATE \quad\textrm{Compute} $\mathrm{d}(\mathcal{Q}_{\rm BC}+\mathcal{R}_{\rm adv})/\mathrm{d}\mathbf{W}$
        \STATE \quad \textrm{Update }$\mathbf{W}$ using \eqref{eqn:gd_w}
        \STATE \textbf{\textit{Stackelberg Adv Reg}: }
        \STATE \quad\textrm{Compute} $\mathrm{d}\mathcal{Q}_{\rm SAR}/\mathrm{d}\mathbf{W}$ using \eqref{eqn:stackelberg_grad}
        \STATE \quad$\mathbf{W}\gets\mathrm{Optimizer}(\mathrm{d}\mathcal{Q}_{\rm{SAR}}/\mathrm{d}\mathbf{W})$ 
        \ENDFOR
    \ENDFOR
    \end{algorithmic}
\end{algorithm}

To simplify the notation, we omit the indices on the trajectory sample points $\mathbf{x}$. We solve
\begin{align}
    & \min_{\mathbf{W}}\mathcal{Q}_{\rm SAR}(\mathbf{W})=\mathcal{Q}_{\rm BC}(\mathbf{X},\mathbf{U},\mathbf{W})+\frac{\alpha}{N}\sum r(\mathbf{x},\mathbf{W},\boldsymbol{\delta}^{K}), \label{eqn:salt_leader}\\
    & \hspace{0.05in}\text{s.t.}\ \boldsymbol{\delta}^{K}(\mathbf{W})=U^K \circ U^{K-1}\circ\cdots\circ U^1(\boldsymbol\delta^{0}).
    \notag
\end{align}
The policy parameter $\mathbf{W}$ in Eq.~{\eqref{eqn:salt_leader}} is the leader, and the perturbation $\boldsymbol \delta(\mathbf{W})$ is the follower. Here,
$\circ$ denotes operator composition, i.e., $f(\cdot) \circ g(\cdot)=f(g(\cdot))$. Each $U^k$ for $k=1,\cdots,K$ represents the $k^{\rm th}$ step update operator for the follower's strategy. The operators are defined by pre-selected optimization algorithms such as stochastic gradient descent (SGD) or Adam~\cite{kingma2014adam}.

In Stackelberg adversarial training, the leader acknowledges the strategy of the follower by treating the perturbations (the follower) as a function of the policy parameters (the leader). Correspondingly, we solve for the policy parameters using gradient descent, where the Stackelberg gradient is
\begin{equation}
    \label{eqn:stackelberg_grad}
    \frac{\rm{d}\mathcal{Q}_{\rm SAR}(\mathbf{W})}{\rm{d}\mathbf{W}}=\underbrace{\frac{\rm{d}\mathcal{Q}_{\rm BC}(\mathbf{X},\mathbf{U},\mathbf{W})}{\rm{d}\mathbf{W}}+\alpha\frac{\partial r(\mathbf{x}, \mathbf{W}, \boldsymbol{\delta}^K)}{\partial \mathbf{W}}}_{\textrm{leader}}+\underbrace{\alpha\frac{\partial r(\mathbf{x}, \mathbf{W}, \boldsymbol{\delta}^K)}{\partial \boldsymbol\delta^K}\frac{\rm{d}\boldsymbol{\delta}^K}{\rm{d}\mathbf{W}}}_{\textrm{leader-follower interaction}}.
\end{equation}

In comparison, the conventional adversarial regularization in Eq.~{\eqref{minmax-pl}} uses only the leader term and does not consider the leader-follower interaction. 

The most expensive term to compute in Eq.~{\eqref{eqn:stackelberg_grad}} is ${\rm{d}\boldsymbol\delta^K}/{\rm{d}\mathbf{W}}$.
Recall that we have $\boldsymbol \delta^k = U^k (\boldsymbol \delta^{k-1})$, where $U^k$ is an update operator, e.g., a one-step gradient ascent. As a short-hand, we write
\begin{equation*}
    \boldsymbol{\delta}^k(\mathbf{W}) = \boldsymbol{\delta}^{k-1}(\mathbf{W}) + \Delta(\mathbf{x}, \boldsymbol{\delta}^{k-1}(\mathbf{W}), \mathbf{W}),
\end{equation*}
where $\Delta(\mathbf{x}, \boldsymbol{\delta}^{k-1}(\mathbf{W}), \mathbf{W})$ signifies the update from $\boldsymbol \delta^{k-1}$ to $\boldsymbol \delta^k$. Then we have
\begin{align*}
    \frac{\rm{d}\boldsymbol{\delta}^k}{\rm{d}\mathbf{W}}=\frac{\rm{d}\boldsymbol{\delta}^{k-1}}{\rm{d}\mathbf{W}}+\frac{\partial \Delta(\mathbf{x}, \boldsymbol{\delta}^{k-1}, \mathbf{W})}{\partial \mathbf{W}} + \frac{\partial \Delta(\mathbf{x}, \boldsymbol{\delta}^{k-1}, \mathbf{W})}{\partial \boldsymbol{\delta}^{k-1}}\frac{\rm{d}\boldsymbol{\delta}^{k-1}}{\rm{d}\mathbf{W}}.
\end{align*}


This recursive differentiation can be efficiently computed using deep learning libraries, such as \textit{PyTorch}~\cite{NEURIPS2019_9015}. Please refer to \cite{zuo2021adversarial} for more details. The overall adversarial regularization algorithm is shown in Algorithm~\ref{pseudo:adv_policy_learning}.

\section{Experiments}

We evaluate VERONICA on cart-pole swing-up and Kuka arm manipulation tasks. The manipulation scenarios are shown in Figure~\ref{fig:tasks}. The experiments are shown in the video\footnote{The link to the video is \url{https://youtu.be/2zlAC9Xs8Bg}.}. We compare smoothness, generalization, and robustness of policies trained with Gaussian perturbations, conventional adversarial regularization (VERONICA-AR), and SAR (VERONICA-SAR). We do not include tangent propagation due to the excessive computational requirements to compute the Jacobian. We also demonstrate that the neural control policy is able to handle simple multi-modal dynamics for the pick and place task.


For Kuka manipulation tasks, the simulation environment is implemented in \textit{PyBullet}~\cite{coumans2021}. We solve for TO described in Eq.~{\eqref{eqn:admm_to}} using DDP implemented in \textit{Crocoddyl}~\cite{mastalli20crocoddyl}. For hopper locomotion tasks, we implement both the simulation environment and a direct TO algorithm in \textit{Drake}~\cite{drake}. The adversarially regularized policy learning algorithm is implemented in \textit{PyTorch}~\cite{NEURIPS2019_9015} and \textit{Higher}~\cite{grefenstette2019generalized}. The implementation details can be found in Appendix \ref{appendix:imp_details}.

\begin{figure}[htb!]
    \centering
    \includegraphics[width=0.5\textwidth]{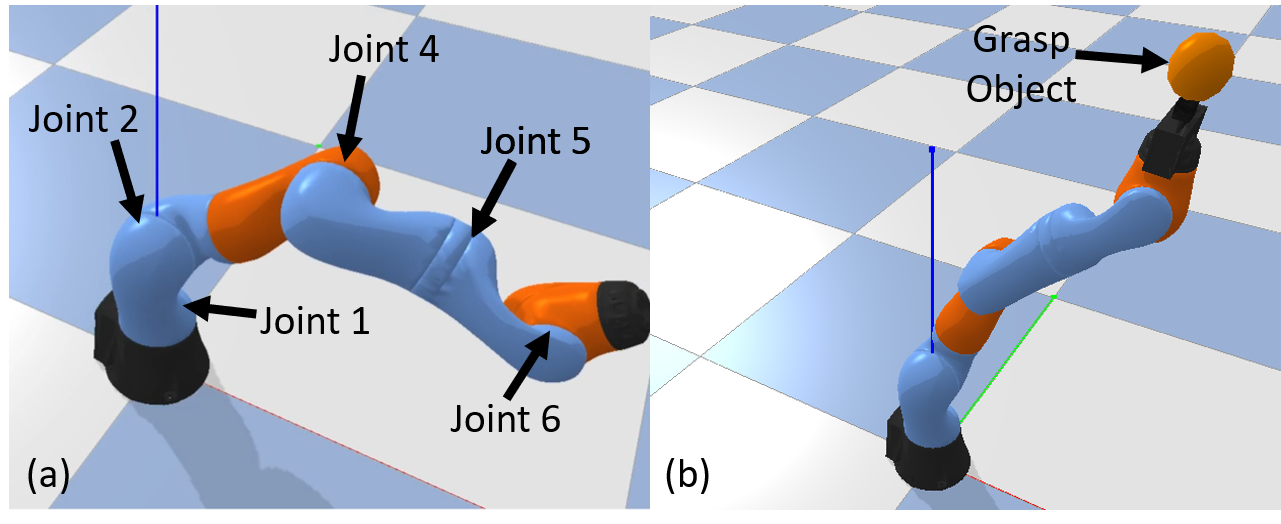}
    \caption{The Kuka arm manipulation scenarios in simulation. (a) Kuka IIWA arm reaching: the learned policy controls the arm to reach a predefined joint configuration. In 3-DOF reaching experiments, only joints 2, 4, and 6 are active degrees-of-freedom (DOFs), making the arm equivalent to a planar manipulator. In 5-DOF experiments, joints 1, 2, 4, 5, and 6 are active DOFs; (b) The Kuka arm pick and place task: an additional object is grasped by the Kuka arm during this task.}
    \label{fig:tasks}
\end{figure}

\textbf{Policy Smoothness: } We qualitatively examine the smoothness of our neural control policy by inspecting a typical policy roll-out for cart-pole swing-up and Kuka arm reaching tasks, as shown in Figure~\ref{fig:smoothness_experiment}. Figure~\ref{fig:smoothness_experiment}(a) shows the smoothness comparison during a cart-pole swing-up. VERONICA produced visually smoother force sequences comparing to Gaussian perturbation. Figure~\ref{fig:smoothness_experiment}(b) displays the torque sequence
of Kuka joint 2 during a reaching task. 
The policy trained by Gaussian perturbation generates a non-smooth torque profile around the initial position of the task, indicating that the Gaussian perturbation is not sufficient to prevent overfitting at the initial phase of the trajectory, where the torque changes relatively quickly with respect to state. 
In comparison, the VERONICA-AR and VERONICA-SAR policies produce smoother control sequences that track the baseline closely. To inspect the smoothness of the neural control policies, we plot the torque output on Kuka joint 2 against the joint angle in Figure~\ref{fig:smoothness_experiment}(c). VERONICA successfully penalize against the non-smooth peak that appeared in the torque profile of the Gaussian perturbed policy.

\begin{figure}[htb!]
    \centering \includegraphics[width=0.92\textwidth]{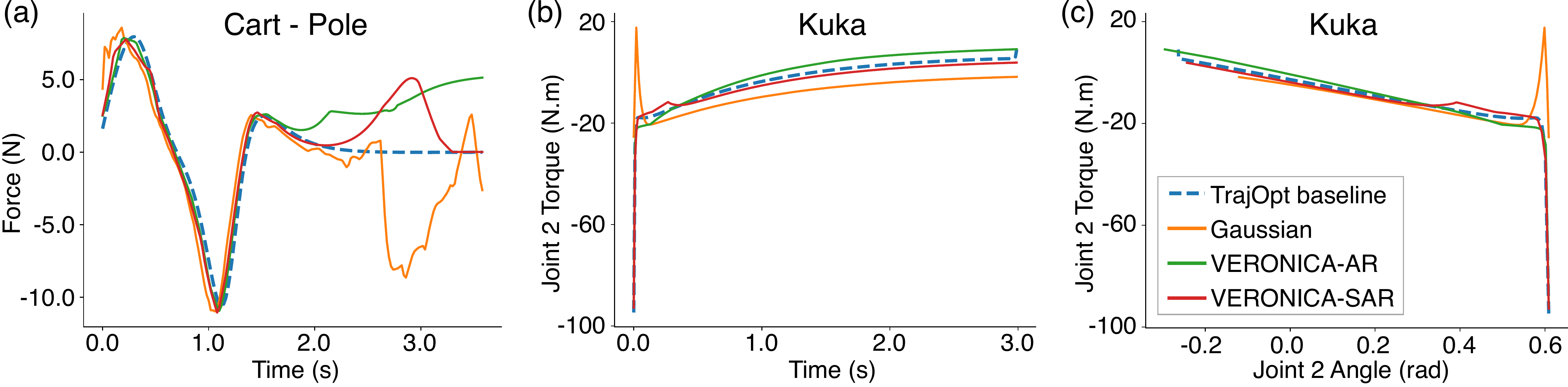}
     \caption{Comparison of control output smoothness for cart-pole and Kuka arm reaching tasks. Trajectory optimization baseline is marked as a dashed line. (a) Time sequence of forces applied onto the cart during swing-up. (b-c) Torque output for Kuka joint 2 with respect to time and joint 2 angle.}
    \label{fig:smoothness_experiment}
\end{figure}

\begin{figure}[htb!]
    \centering
    \includegraphics[width=0.92\textwidth]{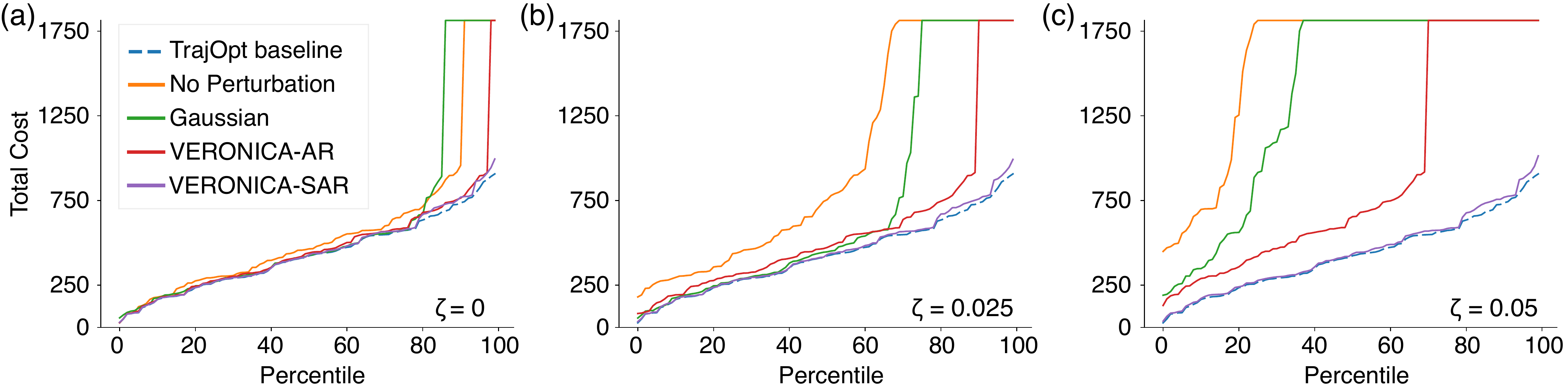}
    \vspace{-0.05in}
    \caption{\captionsize Cost percentile plot for 3-DOF arm reaching task with 100 different initializations and under different disturbances on sensor measurement. Disturbances are drawn from a uniform distribution bounded by $\zeta$. Policies trained with no perturbation, Gaussian perturbation, VERONICA-AR, and VERONICA-SAR are compared against an undisturbed TO baseline. The plot is capped at 2 times the maximum baseline cost. A cost curve that exceeds the plotting cap indicates that a percentage of policy roll-outs lead to unstable robot motion.}
    \label{fig:cum_cost_percentile}
\end{figure}

\textbf{Generalization Performance: } To evaluate the generalization performances of VERONICA, we perform policy roll-outs with 100 different initializations in an undisturbed environment, as seen in Figure~\ref{fig:cum_cost_percentile}(a). The adversarially regularized policies produce lower costs because the policies trained with no perturbation or Gaussian perturbation are unable to generate stable robot motions under some initializations. Figure~\ref{fig:3_DOF_example} displays an example of an arm reaching task that Gaussian perturbation cannot handle. Although a vast majority of roll-outs with the VERONICA-AR policy are stable, a small percentage (2\%) produces unstable robot motions that fail to achieve the task. In comparison, the VERONICA-SAR policy leads to stable and near-optimal robot motions across all attempts, confirming our hypothesis that VERONICA-SAR helps enhance numerical stability comparing to VERONICA-AR.


\begin{figure}[htb!]
\vspace{-0.1in}
    \centering
    \includegraphics[width=0.92\textwidth]{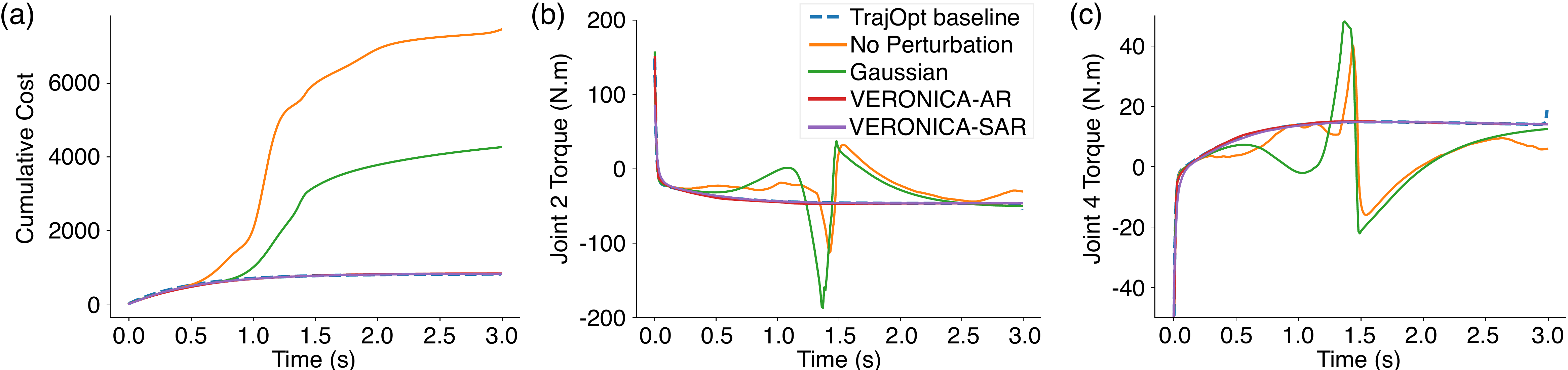}
    \caption{Example of an undisturbed policy roll-out for a 3-DOF manipulator reaching task where Gaussian perturbation fails. The undisturbed TO result is provided as a baseline. (a) Cumulative cost for policy roll-out (b-c) Torque outputs on joints 2 and 4.} 
    \label{fig:3_DOF_example}
    \vspace{-0.1in}
\end{figure}


\textbf{Policy Robustness: }
We evaluate our policies' robustness against three different kinds of disturbances. For sensor noise and environmental uncertainty, we add a uniform noise bounded by an $\ell_{\infty}$-norm ball with radius $\zeta$ onto the sensor measurement and state transition, respectively. As for model mismatch, we modify the URDF file used in policy roll-out by decreasing the mass of each robot link by 0.25 kg. 

We first compare the policies' robustness against different magnitudes of sensor noise, as shown in Figure~\ref{fig:cum_cost_percentile}(b-c). While Gaussian perturbation does provide some robustness comparing to the unregularized policy, VERONICA-AR and VERONICA-SAR consistently outperforms the Gaussian perturbation. Furthermore, VERONICA-AR deviates significantly from the undisturbed TO baseline under a strong sensor noise ($\zeta=0.05$), while VERONICA-SAR remains able to produce stable robot motion and closely track the TO baseline.
\begin{table}[htb!]
\vspace{-0.15in}
\caption{Task Error for 3-DOF Manipulator Reaching Task ($\zeta=0.01$, \textit{Unit: m})}
\label{tab:task_error}
\vspace{-0.06in}
\begin{center}
 \begin{tabular}{c c c c} 
 \hline
  & Gaussian & VERONICA-AR & VERONICA-SAR \\ 
 \hline\hline
 Undisturbed & $1.62\mathrm{e}\text{-}1\pm5.05\mathrm{e}\text{-}2$ & $6.26\mathrm{e}\text{-}2\pm3.96\mathrm{e}\text{-}2$ & $6.39\mathrm{e}\text{-}2\pm2.70\mathrm{e}\text{-}2$ \\ 
 \hline
 Sensor Error &  $1.75\mathrm{e}\text{-}1\pm7.85\mathrm{e}\text{-}2$ & $7.11\mathrm{e}\text{-}2\pm7.57\mathrm{e}\text{-}2$ & $6.61\mathrm{e}\text{-}2\pm2.42\mathrm{e}\text{-}2$ \\
 \hline
 Environment Uncertainty & $1.73\mathrm{e}\text{-}1\pm7.59\mathrm{e}\text{-}2$ & $8.66\mathrm{e}\text{-}2\pm1.03\mathrm{e}\text{-}1$ & $7.75\mathrm{e}\text{-}2\pm3.99\mathrm{e}\text{-}2$ \\
 \hline
 Model Mismatch & $2.14\mathrm{e}\text{-}1\pm8.26\mathrm{e}\text{-}2$ & $5.24\mathrm{e}\text{-}2\pm3.27\mathrm{e}\text{-}2$ & $1.23\mathrm{e}\text{-}1\pm2.16\mathrm{e}\text{-}2$ \\ 
 \hline
\end{tabular}
\end{center}
\vspace{-0.25in}
\end{table}

Table~\ref{tab:task_error} shows the average task errors - the distance between the goal and the actual final positions for the robot arm's end-effector - and their standard deviation for 100 manipulator reaching tasks under different types of disturbances. VERONICA provides significantly lower task errors across all clean and disturbed experiments. Furthermore, VERONICA-SAR leads to a lower standard deviation than VERONICA-AR, indicating that the policy learned by VERONICA-SAR is less prone to outliers comparing to VERONICA-AR.  

\begin{table}
\caption{Median Task Errors for $M$-DOF Manipulator (\textit{Unit: m})}\label{tab:n_dof_task_err}
\begin{center}
\begin{tabular}{c c c}
\hline
 $M=3$ & $M=5$ & $M=7$ \\
\hline\hline
 6.39\rm{e}\text{-}2 & 1.23\rm{e}\text{-}1 & 1.32\rm{e}\text{-}1 \\
\hline
\end{tabular}
\end{center}
\end{table} 
\textbf{Extension to Higher-DOF Manipulators:}
We investigate how the performance of VERONICA-SAR scales to higher state and control dimensions by evaluating the task errors of manipulator reaching tasks for 3, 5, and 7-DOF Kuka arms (Table~\ref{tab:n_dof_task_err}). The task error increases with the dimensionality of the problem, but not significantly. Note that the 5 and 7-DOF experiments involve manipulation in the 3-D space, which lead to much higher problem complexity than the planar 3-DOF Kuka arm configuration, and require larger neural control policies. \ZhigenZ{Figure~\ref{fig:7dof_cum_cost_percentile} indicates that similar to the 3-DOF cases, the proposed Stackelberg adversarial regularization benefits both generalization and robustness performance compared to Gaussian regularization in the 7-DOF Kuka arm reaching tasks.}

\begin{figure}[htb!]
    \centering
    \includegraphics[width=0.92\textwidth]{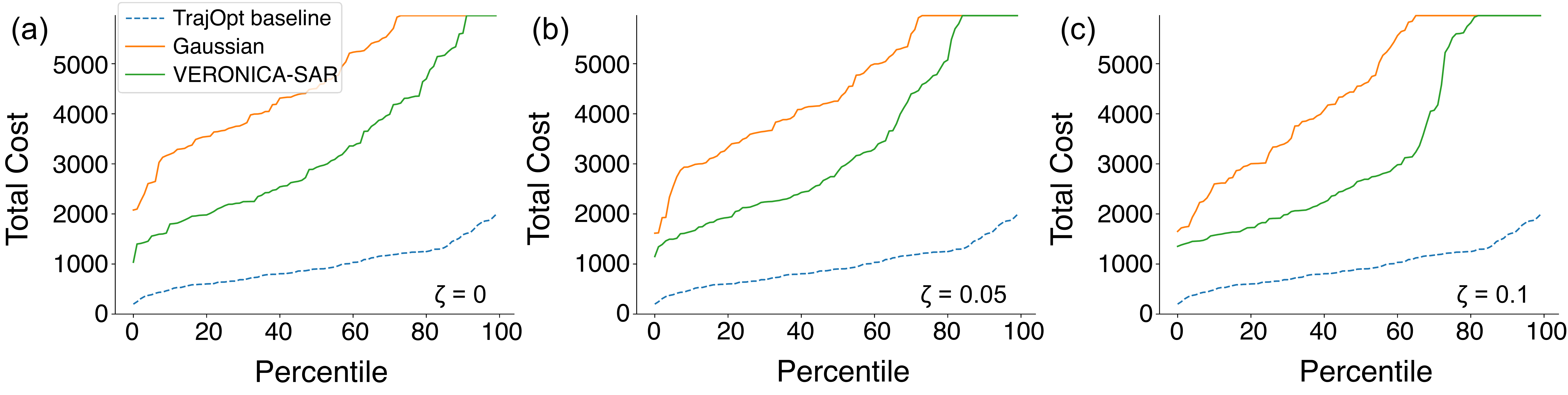}
    \caption{\captionsize \ZhigenZ{Cost percentile plot for 7-DOF arm reaching task with 100 different initializations and under different disturbances on sensor measurements. The plot is capped at 3 times the maximum baseline cost.}}
    \label{fig:7dof_cum_cost_percentile}
\end{figure}

\begin{figure}
\centering
\includegraphics[width=0.4\textwidth]{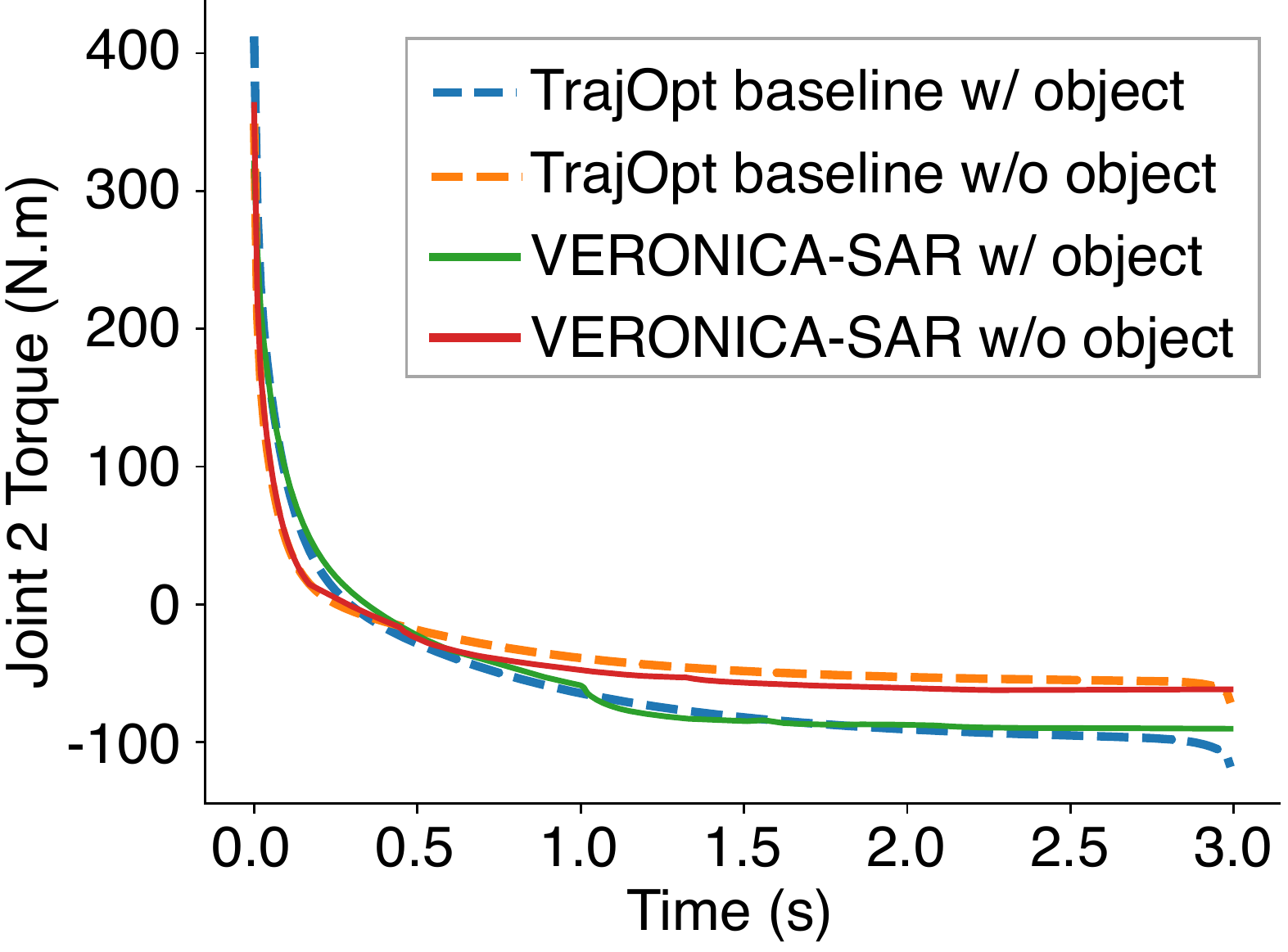}
\caption{Comparison for the control policy outputs with or without grasping a 5kg object.}
\label{fig:pick_and_place_torques}
\end{figure}

\textbf{Preliminary Study of Learning Multimodal Dynamics: } In the pick and place task, we train a network policy to handle the control of the Kuka arm for both free-moving or object-holding scenarios. In order to train the policy applicable for both cases simultaneously, we include a discrete variable in the network input to signify the grasping state of the object. Figure~\ref{fig:pick_and_place_torques} shows the arm's torque output for the same initialization, with or without an object. For simplicity, this experiment assumes that only one object with a known mass, and the object is fixed to a pre-specified position in the gripper when grasped by the arm. In the future, the adaptability of the network policy can be improved by augmenting the input with more information such as the weight of the object and the relative position between the object and the gripper. 

\begin{figure}
\centering
\includegraphics[width=0.4\textwidth]{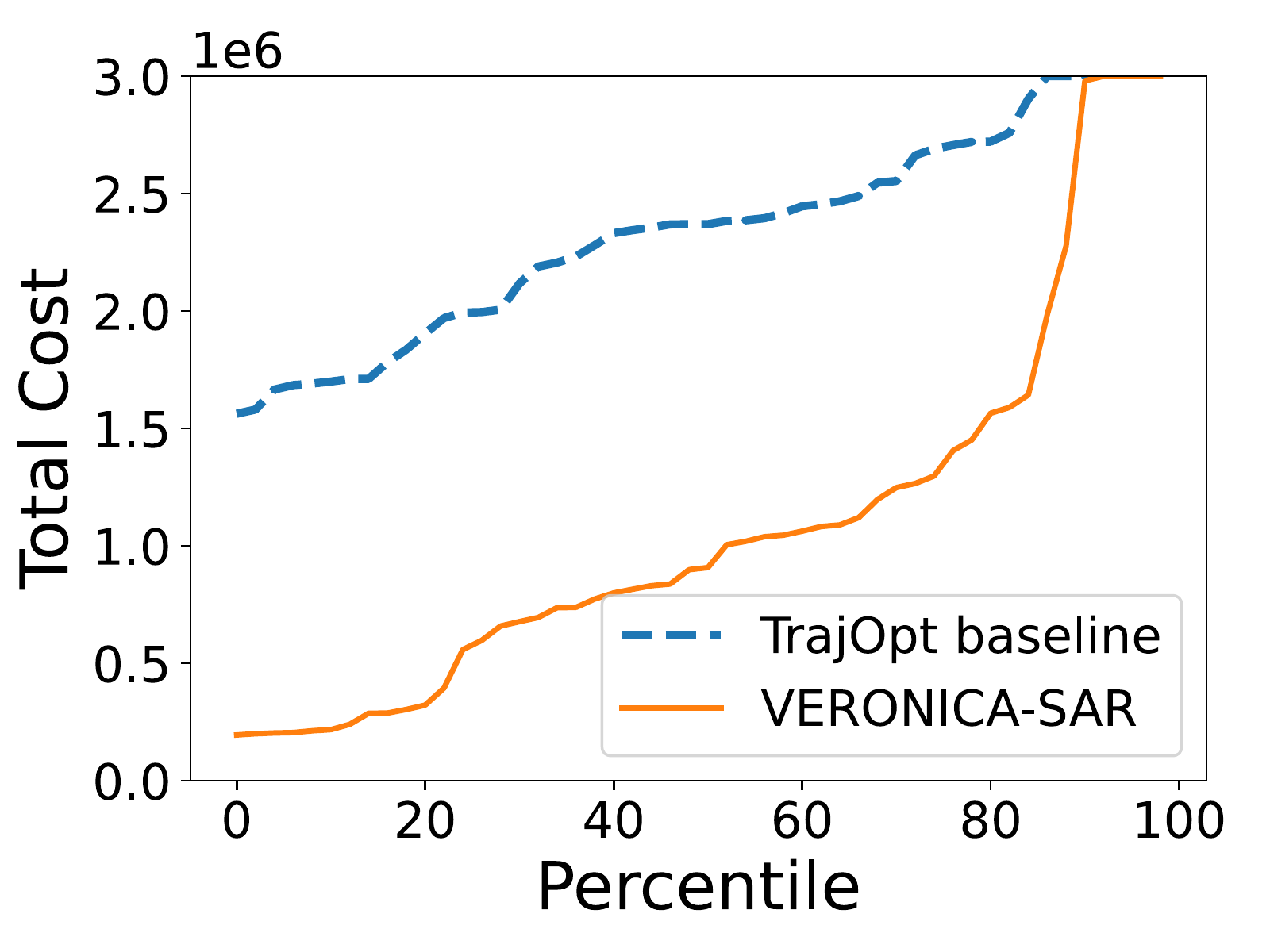}
\caption{Cost percentile plot for hopper locomotion task with 100 different initializations.}
\label{fig:hopper}
\end{figure}

\textbf{Application to Hybrid Locomotion Systems: } We apply VERONICA in hopper locomotion tasks to evaluate the performance of VERONICA in a single leg 5-DOF hopper system, where the  hybrid locomotion trajectories involve intermittent contacts with the terrain. We compare the cost percentile plot between the TO baseline and VERONICA-SAR, as displayed in Figure~\ref{fig:hopper}. Note that the open-loop rollout of trajectories generated by TO baseline performs poorly in simulation due to the model mismatch between TO and simulation environments. In contrast, the policy trained with VERONICA-SAR generates a lower cost hopper motions due to the robustness against model mismatch provided by adversarial perturbation. A visual comparison can be found in the video.

\section{Conclusion}
\label{sec:conclusion}

We present VERONICA, an adversarial regularization framework for combined trajectory optimization and policy learning. We show that the proposed regularizer improves generalization and robustness by enforcing Lipschitz continuity of the policy. Additionally, we propose to further stabilize training by formulating the adversarial regularization as a Stackelberg game. The experiment results in robot manipulation scenarios show that our approach helps to improve the smoothness of the learned policy, which results in a more stable robot motions and lower policy execution costs. Additionally, we demonstrate that policies trained with VERONICA are able to robustly handle various types of disturbances.
Our future work will include various extensions to the proposed framework. For example, we will extend VERONICA to solve more complex manipulation problems involving physical contact and enhance robustness to contact uncertainties. \ZhigenZ{We will employ our method in conjunction with a smoothed contact solver similar to the one in \cite{mordatch2012discovery} to circumvent the discontinuity due to contact phenomena while leveraging the smoothness merit induced by the adversarial regularization. Additionally, adaptive adversarial training, where perturbations are generated by an additional network, can be incorporated to generate variable perturbation radius around contact points.}


Our future work will (i) evaluate the performance of VERONICA in the presence of more types of perturbations and uncertainties, such as varying link moment of inertia and kinematic parameters; (ii) extend VERONICA to solve more complex manipulation and locomotion problems involving physical contact and enhance robustness to contact uncertainties. Adaptive adversarial training, where perturbations are generated by an additional network, can be incorporated to generate variable perturbation radius around contact points.



\clearpage
\printbibliography
\newpage
\appendix

\section*{Supplemental Materials}

\section{Differential Dynamic Programming}
\label{appendix:DDP}

In order to generate each individual trajectory sample satisfying robot rigid body dynamics, we solve the following trajectory optimization (TO) problem formulated as: 
\begin{subequations}\label{eqn:TO}
\begin{align}
    \min_{\mathbf{X}, \mathbf{U}}\quad\mathcal{L}(\mathbf{X}, \mathbf{U})&=\sum_{t=1}^{T-1}\ell(\mathbf{x}^t, \mathbf{u}^t)+\ell_f(\mathbf{x}^T, \mathbf{u}^T) \\
    \text{s.t.}\quad\mathbf{x}^{t+1}&=f(\mathbf{x}^{t}, \mathbf{u}^{t}), 
    \mathbf{x}^0=\mathbf{x}_{\rm{init}}, \\
    \mathbf{X} & \in \mathcal{X}, \ \mathbf{U} \in \mathcal{U},
\end{align}
\end{subequations}

where $\ell(\mathbf{x}^t, \mathbf{u}^t)$ is the cost function at time-step $t$, $\ell_f(\mathbf{x}^T, \mathbf{u}^T)$ represents the terminal trajectory cost at time-step $T$, $\mathbf{x}^{t+1}=f(\mathbf{x}^{t}, \mathbf{u}^{t})$ is the discretized system dynamics, and $\mathcal{X}, \mathcal{U}$ represents additional path constraints on state and control. The running trajectory cost $\ell(\mathbf{x}, \mathbf{u})$ is composed of the a goal tracking term, a control regularization term, and the ADMM residual terms:
\begin{align}\nonumber
    \ell(\mathbf{x}, \mathbf{u})= 
    \mathbf{\hat{x}}^{\top}\mathbf{Q}\mathbf{\hat{x}}+\mathbf{u}^{\top}\mathbf{R}\mathbf{u}+\frac{\rho_x}{2}\|\mathbf{x}-\mathbf{x}^{\rm PL}+\boldsymbol{\lambda}_{\mathbf{x}}\|^2 
    + \frac{\rho_u}{2}\|\mathbf{u}-\mathbf{u}^{\rm PL}+\boldsymbol{\lambda}_{\mathbf{u}}\|^2,
\end{align}
where $\mathbf{\hat{x}}=\mathbf{x}-\mathbf{x}_{\rm goal}$ represents the deviation between the trajectory state $\mathbf{x}$ and goal state $\mathbf{x}_{\rm goal}$ and $\mathbf{Q},\mathbf{R} \succeq 0$ are the weighting matrices for the strength of the regularization. 
The ADMM residual terms $\frac{\rho_x}{2}\|\mathbf{x}-\mathbf{x}^{\rm PL}+\boldsymbol{\lambda}_{\mathbf{x}}\|^2$ and $\frac{\rho_u}{2}\|\mathbf{u}-\mathbf{u}^{\rm PL}+\boldsymbol{\lambda}_{\mathbf{u}}\|^2$ are initialized to be 0 at the first iteration, but eventually have the effect of regularizing the trajectory optimization to be closer to the policy output.

In the following we briefly describe the formulation of DDP, which is used in this work to compute trajectory samples. \cite{jacobson1970differential} provides a detailed representation of DDP in the historical context, and \cite{tassa2014control} presents a control-constrained version of DDP that is widely used in robotics. 

DDP solves the optimization described in Eq.~\eqref{eqn:TO} using a backward pass of Bellman's equation,
\begin{equation}
    V(\mathbf{x}^t) = \min_{\mathbf{u}}[\ell(\mathbf{x}^t, \mathbf{u}^t)+V(\mathbf{x}^{t+1})].
\end{equation}
Let $Q(\delta\mathbf{x}^t, \delta\mathbf{u}^t)$ be the change in local cost function given a perturbation around the $t$th time-step:
\begin{equation}
    Q(\delta\mathbf{x}, \delta\mathbf{u})=\ell(\mathbf{x}+\delta\mathbf{x}, \mathbf{u}+\delta\mathbf{u})-\ell(\mathbf{x}, \mathbf{u})+V(\mathbf{x}+\delta\mathbf{x})-V(\mathbf{x})
\end{equation}
The DDP backward pass computes the second order Taylor expansion of $Q$ and the optimal local perturbation $\delta\mathbf{u}^*$ is given by the local feedback control policy:
\begin{equation}
    \delta\mathbf{u}^*=\mathbf{k}+\mathbf{K}\delta\mathbf{x},
\end{equation}
where $\mathbf{k}=-Q_{uu}^{-1}Q_{u}$ and $\mathbf{K}=-Q_{uu}^{-1}Q_{ux}$. After the backward pass is completed, the DDP forward pass simulates the system by rolling out the system dynamics $\mathbf{x}^{t+1}=f(\mathbf{x}^{t}, \mathbf{u}^{t})$. The backward-forward passes are iterated until convergence.

\section{Algorithm Overview of the Proposed Trajectory Optimization Guided by Adversarially Regularized Policy Learning}
\label{appendix-ADMM}



Algorithm~\ref{pseudo:admm_policy_learning} shows the complete procedure of jointly solving TO and policy learning using ADMM.
\begin{algorithm}[H]
    \caption{TO-Guided Policy Learning Using ADMM}
    \label{pseudo:admm_policy_learning}
    \textbf{Input:} {$P$: total number of ADMM iterations;
    \quad$N$: number of sample trajectories.}
    \begin{algorithmic}
    \STATE $\mathbf{X}_{\rm init}\gets N$ \textrm{trajectory initial conditions}
    \STATE $\boldsymbol{\lambda}_{\mathbf{X}}^{0}, \boldsymbol{\lambda}_{\mathbf{U}}^{0}\gets0$
    
    \FOR{$p=1,\cdots,P$}
    \STATE $\mathbf{X}^{\mathrm{TO}, p},\mathbf{U}^{\mathrm{TO}, p}\gets$ \textrm{compute N trajectories using Eq.~\eqref{eqn:admm_to}} \quad\textbf{(primal TO update)} 
    \STATE $\mathbf{W}^{p}\gets$ \textrm{solve min-max optimization in Eq.~\eqref{eqn:admm_pl} using Algorithm~\ref{pseudo:adv_policy_learning}} \quad\textbf{(policy update)}
    \STATE $\mathbf{X}^{\mathrm{PL}, p},\mathbf{U}^{\mathrm{PL}, p}\gets$ \textrm{optimize using Eq.~(\ref{eqn:xr_update_main})} \quad\textbf{(primal PL update)}
    \STATE $\boldsymbol{\lambda}_{\mathbf{X}}^{p}, \boldsymbol{\lambda}_{\mathbf{U}}^{p}\gets$ \textrm{update using Eq.~(\ref{eqn:dual_updates})} \quad\textbf{(dual update)}
    \ENDFOR
    \RETURN{$\mathbf{W}^P$}
    \end{algorithmic}
\end{algorithm}

\section{Implementation Details}
\label{appendix:imp_details}
We use a fully connected neural network with 2 hidden layers and 8 units per layer for the cart-pole example. The 3-DOF Kuka arm uses 2 hidden layers and 64 units each, the 5-DOF manipulator uses 3 hidden layers and 64 units each, while the 7-DOF manipulator uses a residual network with 3 hidden layers and 256 units. The hopper example, as shown in Figure~\ref{fig:hopper_tasks}, uses a fully connected network with 3 hidden layers and 24 units each.

In all experiments, we train the networks using AdamW~\cite{loshchilov2017decoupled} for policy optimization and stochastic gradient descent (SGD) for adversarial perturbation. The regularization coefficient $\alpha$ is set to 1. The learning rate for the policy learning $\mathrm{lr}_{p}$ is chosen between \{$1\mathrm{e}{\text{-}3}$, $5\mathrm{e}{\text{-}4}$\}, and the learning rate for adversarial perturbation $\mathrm{lr}_{\rm adv}$ is chosen between \{$5\mathrm{e}{\text{-}4}$, $1\mathrm{e}{\text{-}4}$\}. The number of adversarial update steps $K$ is selected from \{1, 3\}, and the adversarial bound $\epsilon$ is chosen from \{$1\mathrm{e}{\text{-}2}$, $5\mathrm{e}{\text{-}3}$\}. The policy is trained for at most 300 epochs, with model averaging in the last 1/4 of total epochs. Also, we apply gradient norm clipping of \{$\infty$, 1\}.

\begin{figure}
    \centering
    \includegraphics[width=0.3\textwidth]{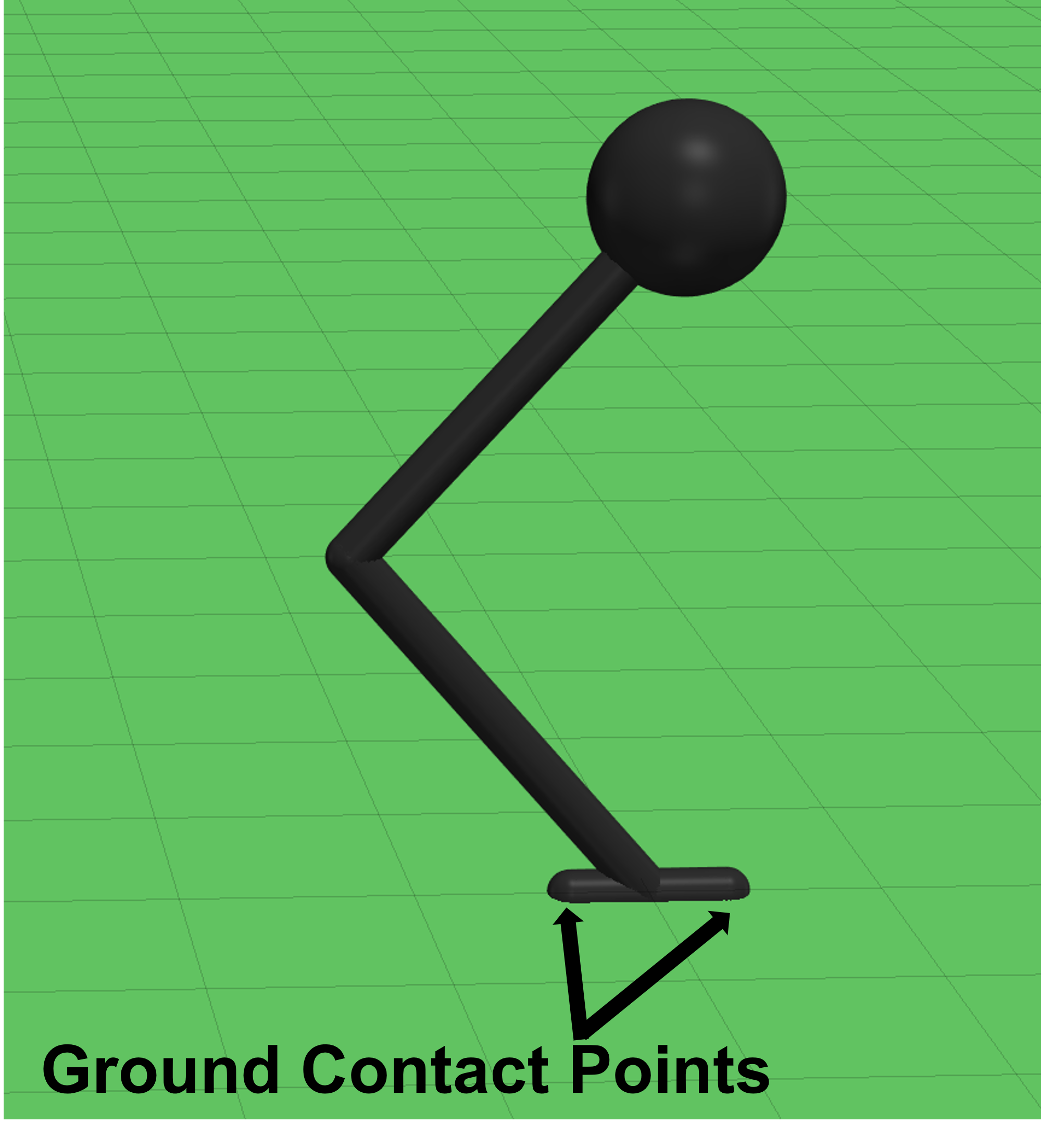}
    \caption{The hopper locomotion tasks in simulation. The single leg hopper has 5 degree-of-freedom, with two contact points with the ground located at the heel and the toe of the hopper.}
    \label{fig:hopper_tasks}
\end{figure}

In ADMM, we apply a trajectory state penalty coefficient $\rho_x$ of \{1, 10, 50\} and a trajectory control penalty coefficient $\rho_{u}$ of 1. We find that the behavioral cloning loss $\mathcal{Q}_{\rm BC}$ decreases over ADMM iterations, but the loss deduction is not significant after 5-10 iterations. Therefore, the ADMM is run until $\mathcal{Q}_{\rm BC}$ stops decreasing, which results in between 5-15 iterations in our experiments. The result for $\mathcal{Q}_{\rm BC}$ plotted with respect to ADMM iterations can be found in Appendix~\ref{appendix:admm_stopping}.

\subsection{3-DOF Kuka Experiments}
We use a fully connected network with 2 hidden layers and 64 units in each layer. The policy input for the reaching task is 9-dimensional, which consists of a 6-dimensional robot state and a 3-dimensional goal configuration. The policy input for the pick and place task is 10-dimensional, with 1 additional input dimension encoding the grasp state. The learning rate for policy parameters $\mathrm{lr}_p$ is set to $1\mathrm{e}\text{-}3$, and the learning rate for adversarial perturbation $\mathrm{lr}_{\rm adv}$ is set to $5\mathrm{e}\text{-}3$. The number of adversarial update step $K$ is selected to be 1, and the adversarial bound $\epsilon$ is $5\mathrm{e}\text{-}3$. We use $N=5000$ trajectory samples with 300 timesteps each. The policy is trained for 300 epochs, with model averaging in the last 75 epochs.

\ZhigenZ{The PPO algorithm as compared in Figure \ref{fig:PPO} is implemented using \textit{Stable Baseline 3}~\cite{stable-baselines3}.}

\subsection{7-DOF Kuka Experiment}
We use a residual network with 3 hidden layers (Figure~\ref{fig:res_net}) to learn the neural control policy for the 7-DOF Kuka experiment. The policy input is 21-dimensional, which consists of a 14-dimensional robot state and a 7-dimensional target joint angles. The learning rate for policy parameters $\mathrm{lr}_p$ is set to $1\mathrm{e}\text{-}3$, and the learning rate for adversarial perturbation $\mathrm{lr}_{\rm adv}$ is set to $1\mathrm{e}\text{-}4$. The number of adversarial update step $K$ is selected to be 1, and the adversarial bound $\epsilon$ is $5\mathrm{e}\text{-}3$. We apply a gradient norm clipping of 1. We use $N=25000$ trajectory samples with 200 timesteps each. The policy is trained for 100 epochs, with model averaging in the last 25 epochs.

\begin{figure}[t]
    \centering
    \includegraphics[width=0.75\textwidth]{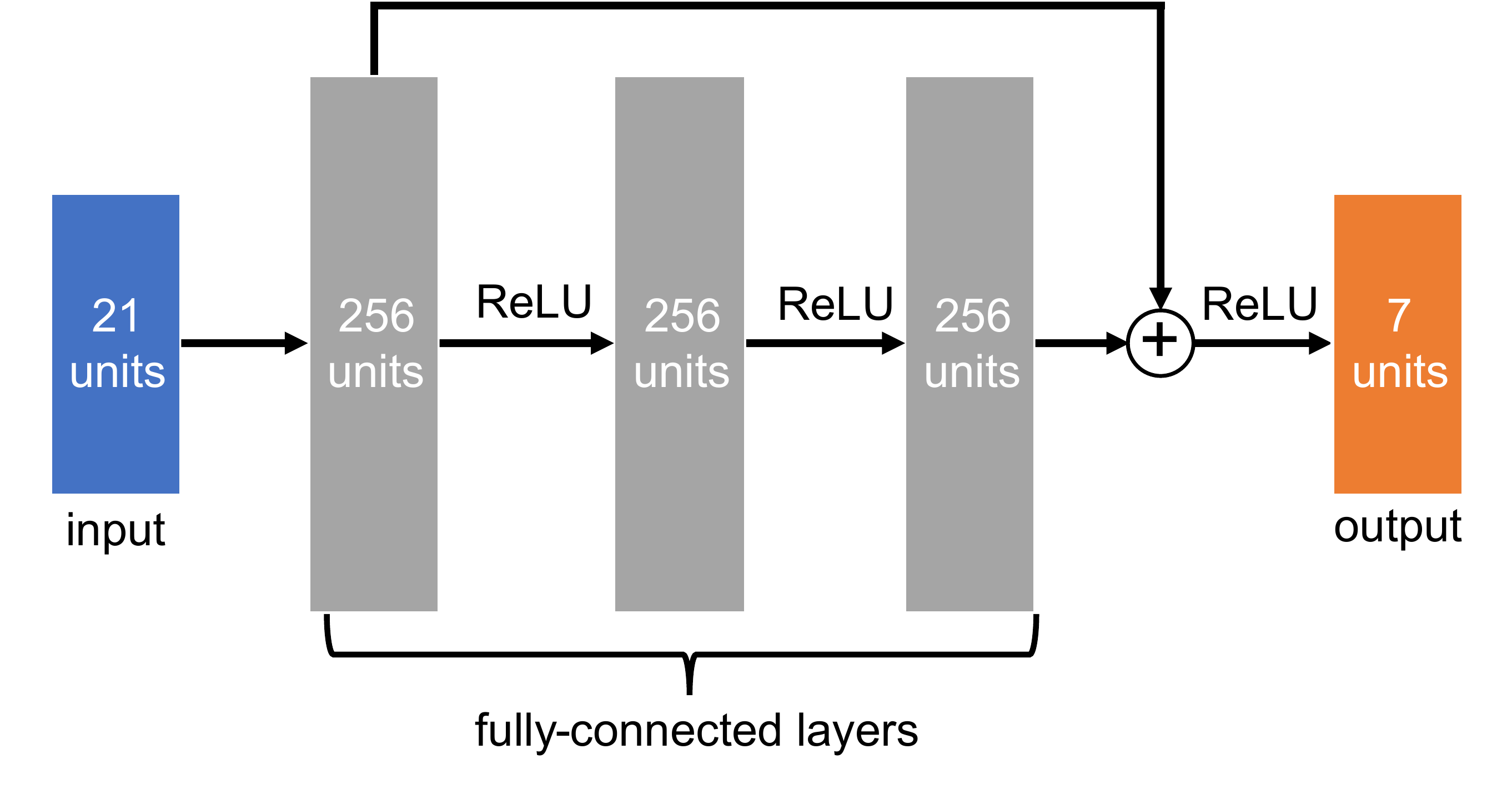}
    \caption{Illustration of the residual network used for 7-DOF Kuka manipulator experiments. The network consists of 3 hidden layers with 256 units each. A skip connection is included from the output of the $1^{\rm st}$ hidden layer to the output of the $3^{\rm rd}$ hidden layer.}
    \label{fig:res_net}
\end{figure}

\section{Policy Behavioral Cloning Loss Over ADMM Iterations}
\label{appendix:admm_stopping}
Figure~\ref{fig:admm_convergence} shows the behavioral cloning loss $\mathcal{Q}_{\rm BC}$ plotted against ADMM iterations. In the cart-pole experiment shown in Figure~\ref{fig:admm_convergence}(a), $\mathcal{Q}_{\rm BC}$ decreases in the first 15 iterations, and gradually increases afterwards. In Kuka experiment (Figure~\ref{fig:admm_convergence}(b)), $\mathcal{Q}_{\rm BC}$ is improved significantly in the first 2 iterations, then only slowly decreases from the $3^{\rm rd}$ iteration onward.
\begin{figure}
    \centering
    \includegraphics[width=0.8\textwidth]{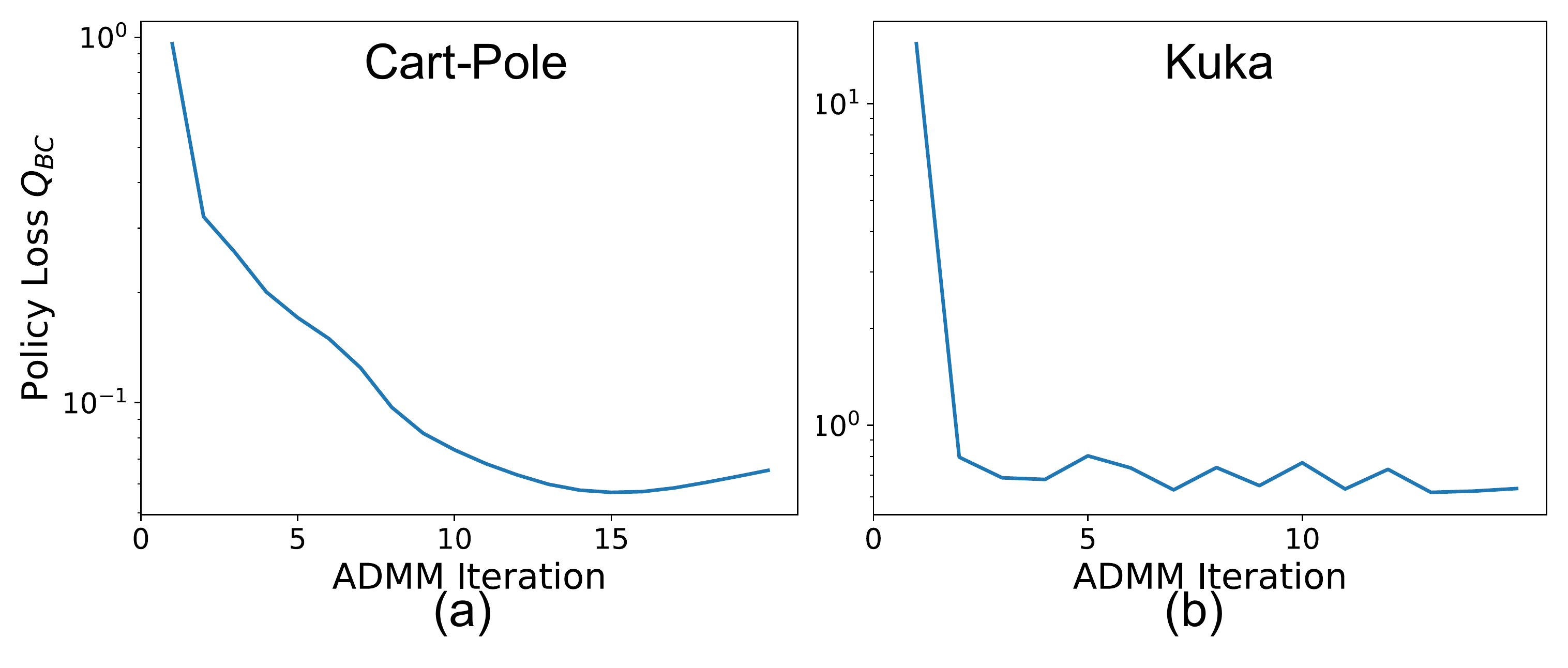}
    \caption{The behavioral cloning losses $\mathcal{Q}_{\rm BC}$ with respect to ADMM iterations. The policies are trained with VERONICA-SAR for (a) cart-pole and (b) 3-DOF Kuka manipulator.}
    \label{fig:admm_convergence}
\end{figure}

\section{Effects of Adversarial Perturbation Bound Value}
We evaluate the effect of adversarial perturbation bound $\epsilon$ by comparing the 3-DOF Kuka arm policies trained by VERONICA-SAR with a set of perturbation values $\epsilon\in\{0, 0.005, 0.01, 0.025, 0.05\}$. As seen in Figure~\ref{fig:cum_cost_percentile_eps}, $\epsilon\in\{0.005, 0.01\}$ provides the best performances and closely track the TO baseline. $\epsilon=0$ is equivalent to the policy trained without perturbation, which does not enjoy the generalization and robustness gains provided by VERONICA. In contrast, the policy performance decreases significantly when $\epsilon>0.025$, indicating that the adversarial perturbation is too strong and causes underfitting.

\begin{figure}[t]
    \centering
    \includegraphics[width=0.5\textwidth]{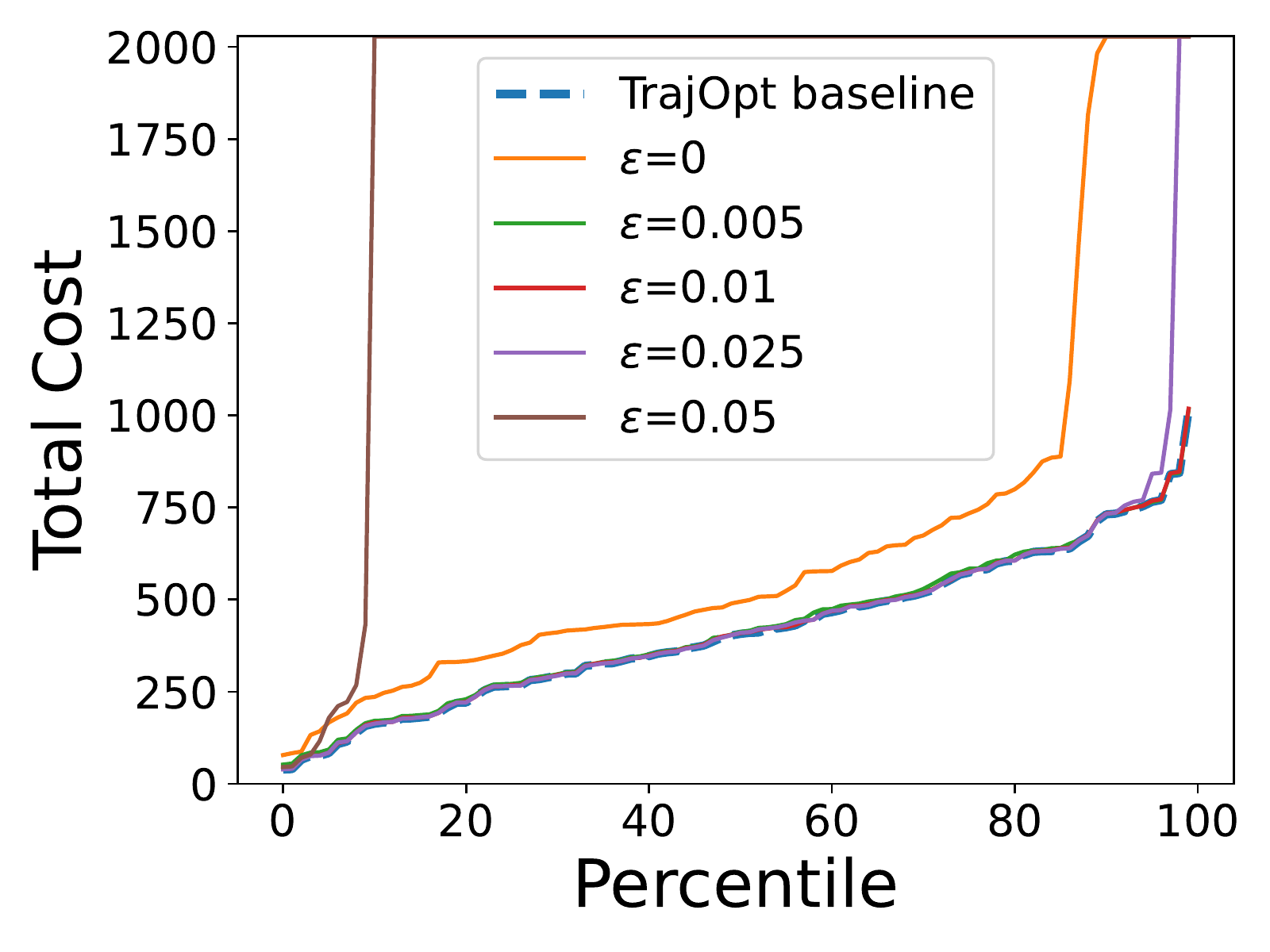}
    \caption{Cost percentile plot for 3-DOF arm policy rollout with 100 different initializations. The policies are trained with different adversarial perturbation bounds $\epsilon$}.
    \label{fig:cum_cost_percentile_eps}
\end{figure}

\ZhigenZ{
\section{Theoretical Analysis on Policy Smoothness and Robustness}
\label{supp_sec:robustness}
In this section, we provide a theoretical analysis on how the Lipschitz continuity improves a neural control policy's robustness. We evaluate the policy's robustness against state disturbances via value discrepancy propagation analysis \cite{xu2019value}, where the policy robustness is analyzed by studying how the error caused by state disturbance propagates in the value functions of the policy. As shown in Appendix~\ref{appendix:value_discrepancy}, the upper bound of the policy robustness (measured by value function discrepancy) is proportional to the Lipschitz constant of the policy. Therefore, controlling the Lipschitz continuity of the policy helps to improve its robustness.}

\ZhigenZ{
We make the assumption that the poilcy $\pi$, the cost function $\ell(\mathbf{x}, \mathbf{u})$, and the system dynamics $f(\mathbf{x}, \mathbf{u})$ are globally Lipschitz continuous. Although these assumptions might not hold in all practical cases, the following discussion provides some insight and intuition about why controlling the smoothness of the policy enhances its robustness against various disturbances.
}
\ZhigenZ{
\subsection{Definitions}
$\pi(\cdot|\mathbf{W})$ denotes a neural control policy with network parameters $\mathbf{W}$. For notation simplicity, $\mathbf{W}$ are omitted in the following discussion. Let $\ell_{\pi}(\mathbf{x}^{(t)})=\ell(\mathbf{x}^{(t)}, \pi(\mathbf{x}^{(t)}))$ denote the cost for policy $\pi$ at state $\mathbf{x}^{(t)}$ on time-step $t$. Similarly, $f_{\pi}(\mathbf{x}^{(t)})=f(\mathbf{x}^{(t)}, \pi(\mathbf{x}^{(t)}))$ represents the system dynamics under policy $\pi$ at state $\mathbf{x}^{(t)}$.
}
\ZhigenZ{
We define the value function $J_{\pi}$ of policy $\pi(\mathbf{x})$ to be the infinite horizon cost with a discount factor $\gamma\in(0,1)$,
\begin{equation}\nonumber
    J_{\pi}(\mathbf{x}^{(0)})= \sum_{t=0}^{\infty} \gamma^{t} \ell_{\pi}(\mathbf{x}^{(t)}).
\end{equation}
We consider the discount factor for convenience of analysis. The results can be extended to the average cost setting, but will be more involved.
}

\ZhigenZ{
The Lipschitz constant of $\pi$, $\ell_{\pi}$, $f_{\pi}$, and $J_{\pi}$ are denoted as $C_{\pi}$, $C_{\ell_{\pi}}$, $C_{f_{\pi}}$, and $C_{J_{\pi}}$ respectively. $C_{\ell}^{\mathbf{u}}$ and $C_{f}^{\mathbf{u}}$ represents the Lipschitz constant of $\ell(\mathbf{x}, \mathbf{u})$ and $f(\mathbf{x}, \mathbf{u})$ with respect to $\mathbf{u}$.
}
\ZhigenZ{
\subsection{Lipschitz Continuity of Value Function}
\textit{
\textbf{Lemma 1:} Given a neural control policy $\pi$ with Lipschitz continuous cost function $\ell_{\pi}$ and dynamics $f_{\pi}$, and let $\gamma C_{f_{\pi}}<1$. The value function $J_{\pi}$ is Lipschitz continuous and the Lipschitz constant is $C_{J_{\pi}} = \frac{C_{\ell_{\pi}}}{1-(\gamma C_{f_{\pi}})^t}$.
}
}

\ZhigenZ{
Proof:
\begin{align}\nonumber
    &\quad\|J_{\pi}(\mathbf{x}^{(0)})-J_{\pi}(\mathbf{y}^{(0)})\|\\\nonumber
    &= \sum_{t=0}^{\infty}\gamma^t\|\ell_{\pi}(f_{\pi}(\mathbf{x}^{(t)}))-\ell_{\pi}(f_{\pi}(\mathbf{y}^{(t)}))\|\\\nonumber
    &\leq \sum_{t=0}^{\infty}C_{\ell_{\pi}} \gamma^t \|f_{\pi}(\mathbf{x}^{(0)})-f_{\pi}(\mathbf{y}^{(0)})\|\\\nonumber
    &\leq (\sum_{t=0}^{\infty}(\gamma C_{f_{\pi}})^t)C_{\ell_{\pi}}\|\mathbf{x}^{(0)}-\mathbf{y}^{(0)}\|\\
    &= \frac{C_{\ell_{\pi}}}{1-(\gamma C_{f_{\pi}})^t}\|\mathbf{x}^{(0)}-\mathbf{y}^{(0)}\|\nonumber
\end{align}
}

\ZhigenZ{
\subsection{Value Discrepancy Under State Disturbances}
\label{appendix:value_discrepancy}
 \textit{\textbf{Lemma 2}} below shows that the value discrepancy for a policy $\pi$ caused by a norm bounded perturbation is proportional to the Lipschitz constant of the policy. 
}

\ZhigenZ{
\textit{\textbf{Lemma 2:} Given a neural control policy $\pi$ and let $\boldsymbol{\delta}^{(t)}$ be the state disturbance at time-step $t$ norm bounded by $\|\boldsymbol{\delta}^{(t)}\|\leq\zeta$. Let $\pi'(\mathbf{x}^{(t)})=\pi(\mathbf{x}^{(t)}+\boldsymbol{\delta}^{(t)})$ denote the disturbed neural control policy. The discrepancy between value functions $J_{\pi'}$ and $J_{\pi}$ has an upper bound of $C_{\pi}(\frac{C_{\ell}^{u}+\gamma C_{J_{\pi}}C_f^u}{1-\gamma})\zeta$.} 
}

\ZhigenZ{
Proof:
}

\ZhigenZ{
The value function $J_{\pi}$ satisfies:
\begin{equation}\nonumber
    J_{\pi}(\mathbf{x}) = \ell_{\pi}(\mathbf{x}) + \gamma J_{\pi}(f_{\pi}(\mathbf{x})).
\end{equation}
Therefore, the value discrepancy due to disturbances $\boldsymbol{\delta}$ can be written as the following:
\begin{align}\nonumber
    &\quad J_{\pi'}(\mathbf{x})-J_{\pi}(\mathbf{x})\\\nonumber
    &= \ell_{\pi'}(\mathbf{x})-\ell_{\pi}(\mathbf{x})+\gamma(J_{\pi'}(f_{\pi'}(\mathbf{x}))-J_{\pi}(f_{\pi}(\mathbf{x})))\\\nonumber
    &\leq C_{\ell}^{\mathbf{u}}\|\pi'(\mathbf{x})-\pi(\mathbf{x})\|+\gamma(J_{\pi}(f_{\pi'}(\mathbf{x}))-J_{\pi}(f_{\pi}(\mathbf{x})))+\gamma(J_{\pi'}(f_{\pi'}(\mathbf{x}))-J_{\pi}(f_{\pi'}(\mathbf{x})))\\\nonumber
    &\leq C_{\ell}^{\mathbf{u}}C_{\pi}\zeta+\gamma C_{J_{\pi}} C_f^{\mathbf{u}} C_{\pi}\zeta+\gamma(J_{\pi'}(f_{\pi'}(\mathbf{x}))-J_{\pi}(f_{\pi'}(\mathbf{x})))\quad\quad\quad\quad\textrm{(\textbf{by \textit{Lemma 1}})}\\\nonumber
    &\leq C_{\pi}(\frac{C_{\ell}^{\mathbf{u}}+\gamma C_{J_{\pi}}C_f^{\mathbf{u}}}{1-\gamma})\zeta.
\end{align}
}
\end{document}